\documentclass[10pt,twocolumn,letterpaper]{article}

\usepackage{cvpr}
\usepackage{times}
\usepackage{epsfig}
\usepackage{graphicx}
\usepackage{amsmath}
\usepackage{amssymb}
\usepackage{booktabs}
\usepackage{lipsum}

\usepackage[pagebackref=true,breaklinks=true,letterpaper=true,colorlinks,bookmarks=false]{hyperref}

\cvprfinalcopy % *** Uncomment this line for the final submission

\def\cvprPaperID{642} % *** Enter the CVPR Paper ID here

\ifcvprfinal\fi
\begin{document}

%%%%%%%%% TITLE

\title{ShuffleNet: An Extremely Efficient Convolutional Neural Network for Mobile Devices}

\author{
	\begin{tabular*}{0.7\textwidth}{@{\extracolsep{\fill}}cccc}
		Xiangyu Zhang\footnotemark[1] & Xinyu Zhou\footnotemark[1] & Mengxiao Lin & Jian Sun
	\end{tabular*}
\\
Megvii Inc (Face++) \\
{\tt\small \{zhangxiangyu,zxy,linmengxiao,sunjian\}@megvii.com}
}

\maketitle

\newcommand\blfootnote[1]{% 
	\begingroup 
	\renewcommand\thefootnote{}\footnote{#1}% 
	\addtocounter{footnote}{-1}% 
	\endgroup 
}
\blfootnote{* Equally contribution.}

%%%%%%%%% ABSTRACT
\begin{abstract}
We introduce an extremely computation-efficient CNN architecture named ShuffleNet, which is designed specially for mobile devices with very limited computing power (e.g., 10-150 MFLOPs). The new architecture utilizes two new operations, pointwise group convolution and channel shuffle, to greatly reduce computation cost while maintaining accuracy. Experiments on ImageNet classification and MS COCO object detection demonstrate the superior performance of ShuffleNet over other structures, e.g. lower top-1 error (absolute 7.8\%) than recent MobileNet~\cite{howard2017mobilenets} on ImageNet classification task, under the computation budget of 40 MFLOPs. On an ARM-based mobile device, ShuffleNet achieves $\sim$13$\times$ actual speedup over AlexNet while maintaining comparable accuracy.  
\end{abstract}

%%%%%%%%% BODY TEXT
\section{Introduction}

Building deeper and larger convolutional neural networks (CNNs) is a primary trend for solving major visual recognition tasks~\cite{krizhevsky2012imagenet,he2016deep,szegedy2015going,girshick2014rich,ren2015faster,long2015fully}. The most accurate CNNs usually have hundreds of layers and thousands of channels~\cite{he2016deep,szegedy2016rethinking,szegedy2016inception,xie2016aggregated}, thus requiring computation at billions of FLOPs. This report examines the opposite extreme:  pursuing the best accuracy in very limited computational budgets at tens or hundreds of MFLOPs, focusing on common mobile platforms such as drones, robots, and smartphones.  Note that many existing works~\cite{{jaderberg2014speeding,lebedev2014speeding,zhang2015efficient,zhang2016accelerating,wen2016learning,rastegari2016xnor}} focus on pruning, compressing, or low-bit representing a ``basic" network architecture. Here we aim to explore a highly efficient basic architecture specially designed for our desired computing ranges. 

We notice that state-of-the-art basic architectures such as \emph{Xception}~\cite{chollet2016xception} and \emph{ResNeXt}~\cite{xie2016aggregated} become less efficient in extremely small networks because of the costly dense $1\times 1$ convolutions. We propose using \emph{pointwise group convolutions} to reduce computation complexity of $1\times 1$ convolutions. To overcome the side effects brought by group convolutions, we come up with a novel \emph{channel shuffle} operation to help the information flowing across feature channels. Based on the two techniques, we build a highly efficient architecture called \emph{ShuffleNet}. Compared with popular structures like ~\cite{simonyan2014very,he2016deep,xie2016aggregated}, for a given computation complexity budget, our ShuffleNet allows more feature map channels, which helps to encode more information and is especially critical to the performance of very small networks.

We evaluate our models on the challenging ImageNet classification~\cite{deng2009imagenet,russakovsky2015imagenet} and MS COCO object detection~\cite{lin2014microsoft} tasks. A series of controlled experiments shows the effectiveness of our design principles and the better performance over other structures. Compared with the state-of-the-art architecture \emph{MobileNet}~\cite{howard2017mobilenets}, ShuffleNet achieves superior performance by a significant margin, e.g. absolute 7.8\% lower ImageNet top-1 error at level of 40 MFLOPs. 

We also examine the speedup on real hardware, i.e. an off-the-shelf ARM-based computing core. The ShuffleNet model achieves $\sim$\textbf{13$\times$} \emph{actual} speedup (theoretical speedup is 18$\times$) over AlexNet~\cite{krizhevsky2012imagenet} while maintaining comparable accuracy.

\begin{figure*}[ht]
	\centering
	\includegraphics[height=2.3in]{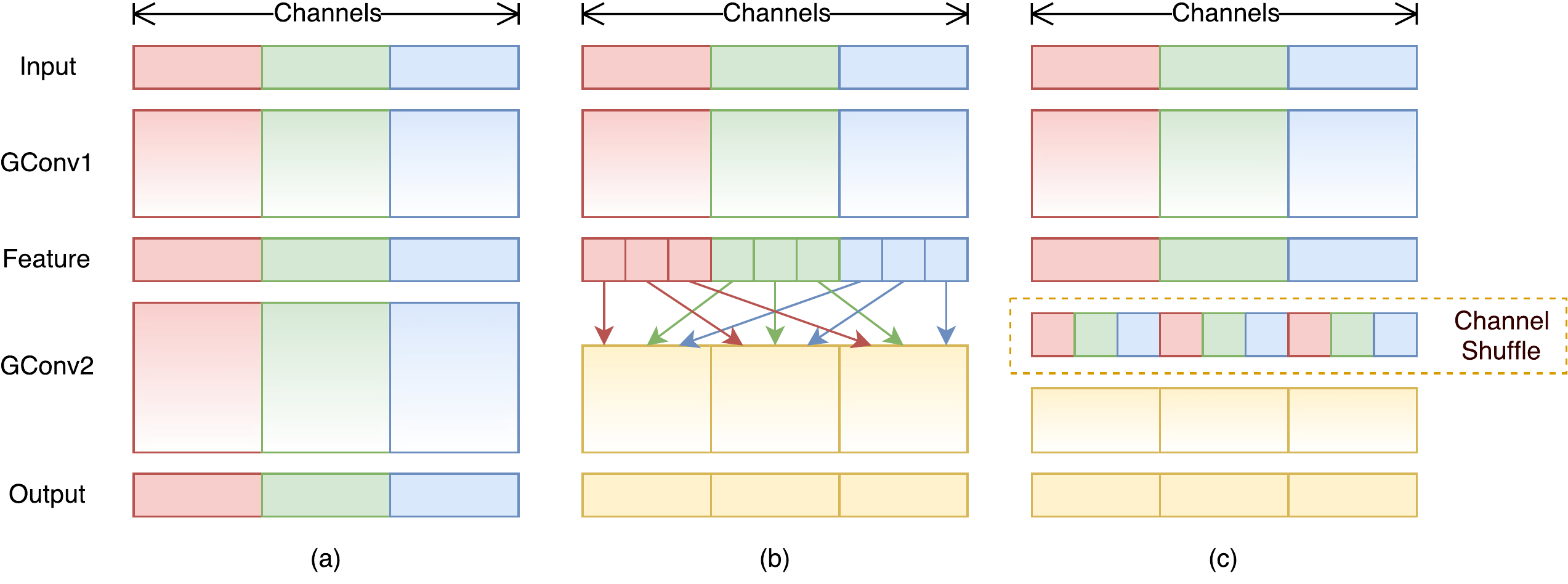}
	\caption{Channel shuffle with two stacked group convolutions. GConv stands for group convolution. a) two stacked convolution layers with the same number of groups. Each output channel only relates to the input channels within the group. No cross talk; b) input and output channels are fully related when GConv2 takes data from different groups after GConv1; c) an equivalent implementation to b) using channel shuffle.}
	\label{fig:channelshuffle}
\end{figure*}

\section{Related Work}

\paragraph{Efficient Model Designs}
The last few years have seen the success of deep neural networks in computer vision tasks~\cite{krizhevsky2012imagenet,vinyals2015show,ren2015faster}, in which model designs play an important role. The increasing needs of running high quality deep neural networks on embedded devices encourage the study on efficient model designs~\cite{he2015convolutional}.
For example, \emph{GoogLeNet}~\cite{szegedy2015going} increases the depth of networks with much lower complexity compared to simply stacking convolution layers. \emph{SqueezeNet}~\cite{iandola2016squeezenet} reduces parameters and computation significantly while maintaining accuracy. \emph{ResNet}~\cite{he2016deep,he2016identity} utilizes the efficient bottleneck structure to achieve impressive performance. 
\emph{SENet}~\cite{hu2017squeeze} introduces an architectural unit that boosts performance at slight computation cost.
Concurrent with us, a very recent work~\cite{zoph2017learning} employs reinforcement learning and model search to explore efficient model designs. The proposed mobile \emph{NASNet} model achieves comparable performance with our counterpart ShuffleNet model (26.0\% @ 564 MFLOPs vs. 26.3\% @ 524 MFLOPs for ImageNet classification error). But~\cite{zoph2017learning} do not report results on extremely tiny models (e.g. complexity less than 150 MFLOPs), nor evaluate the actual inference time on mobile devices.

\paragraph{Group Convolution}
The concept of group convolution, which was first introduced in \emph{AlexNet}~\cite{krizhevsky2012imagenet} for distributing the model over two GPUs, has been well demonstrated its effectiveness in ResNeXt~\cite{xie2016aggregated}. Depthwise separable convolution proposed in Xception~\cite{chollet2016xception} generalizes the ideas of separable convolutions in Inception series~\cite{szegedy2016rethinking,szegedy2016inception}. Recently, MobileNet~\cite{howard2017mobilenets} utilizes the depthwise separable convolutions and gains state-of-the-art results among lightweight models. Our work generalizes group convolution and depthwise separable convolution in a novel form.

\paragraph{Channel Shuffle Operation}
To the best of our knowledge, the idea of channel shuffle operation is rarely mentioned in previous work on efficient model design, although CNN library \emph{cuda-convnet}~\cite{krizhevsky2012cuda} supports ``random sparse convolution" layer, which is equivalent to random channel shuffle followed by a group convolutional layer. Such ``random shuffle" operation has different purpose and been seldom exploited later. Very recently, another concurrent work ~\cite{zhang2017interleaved} also adopt this idea for a two-stage convolution. However, ~\cite{zhang2017interleaved} did not specially investigate the effectiveness of channel shuffle itself and its usage in tiny model design.

\paragraph{Model Acceleration} 
This direction aims to accelerate inference while preserving accuracy of a pre-trained model.
Pruning network connections~\cite{han2015deep, han2015learning} or channels~\cite{wen2016learning} reduces 
redundant connections in a pre-trained model while maintaining performance.
Quantization~\cite{soudry2014expectation, rastegari2016xnor, wu2016quantized, zhou2016dorefa, zhou2017incremental} and 
factorization~\cite{lebedev2014speeding, jaderberg2014speeding, jin2014flattened, wang2016factorized} are proposed in
literature to reduce redundancy in calculations to speed up inference.
Without modifying the parameters, optimized convolution algorithms implemented by FFT~\cite{mathieu2013fast,vasilache2014fast} and other methods~\cite{bagherinezhad2016lcnn} decrease time consumption in practice.
Distilling~\cite{hinton2015distilling} transfers knowledge from
large models into small ones, which makes training small models easier. 

\section{Approach}

\begin{figure*}[ht]
	\centering
	\includegraphics[height=2.8in]{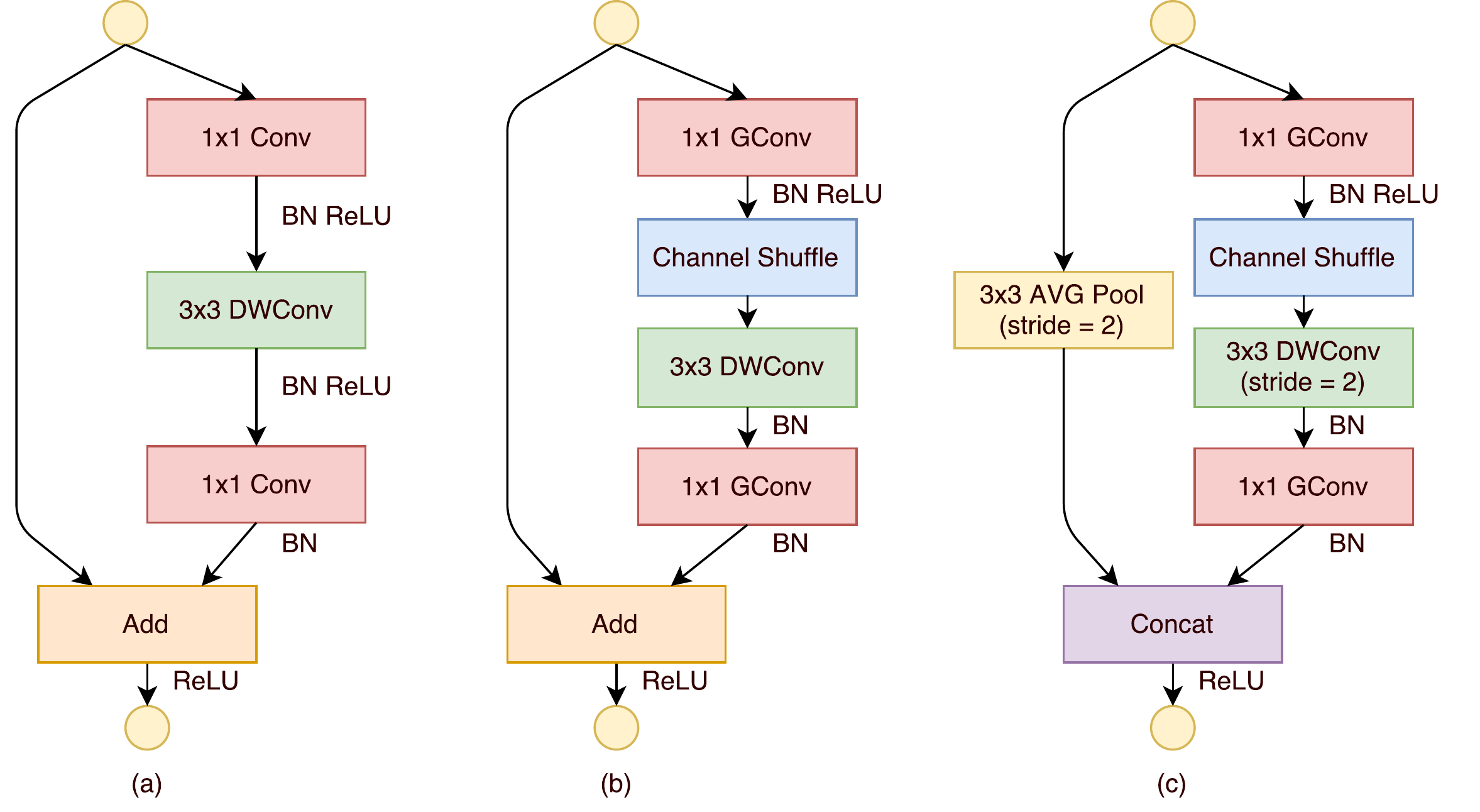}
	\caption{ShuffleNet Units. a) bottleneck unit~\cite{he2016deep} with depthwise convolution (DWConv)~\cite{chollet2016xception,howard2017mobilenets}; b) ShuffleNet unit with pointwise group convolution (GConv) and channel shuffle; c) ShuffleNet unit with stride = 2.}
	\label{fig:unit}
\end{figure*}

\subsection{Channel Shuffle for Group Convolutions}
\label{sec:channelshuffle}

Modern convolutional neural networks~\cite{simonyan2014very,szegedy2015going,szegedy2016rethinking,szegedy2016inception,he2016deep,he2016identity} usually consist of repeated building blocks with the same structure. Among them, state-of-the-art networks such as \emph{Xception}~\cite{chollet2016xception} and \emph{ResNeXt}~\cite{xie2016aggregated} introduce efficient depthwise separable convolutions or group convolutions into the building blocks to strike an excellent trade-off between representation capability and computational cost. However, we notice that both designs do not fully take the $1\times1$ convolutions (also called \emph{pointwise convolutions} in ~\cite{howard2017mobilenets}) into account, which require considerable complexity. For example, in ResNeXt~\cite{xie2016aggregated} only $3\times3$ layers are equipped with group convolutions. As a result, for each residual unit in ResNeXt the pointwise convolutions occupy 93.4\% multiplication-adds (cardinality = 32 as suggested in ~\cite{xie2016aggregated}). In tiny networks, expensive pointwise convolutions result in limited number of channels to meet the complexity constraint, which might significantly damage the accuracy.

To address the issue, a straightforward solution is to apply channel sparse connections, for example group convolutions, also on $1\times1$ layers. By ensuring that each convolution operates only on the corresponding input channel group, group convolution significantly reduces computation cost. However, if multiple group convolutions stack together, there is one side effect: outputs from a certain channel are only derived from a small fraction of input channels. Fig~\ref{fig:channelshuffle} (a) illustrates a situation of two stacked group convolution layers. It is clear that outputs from a certain group only relate to the inputs within the group. This property blocks information flow between channel groups and weakens representation.

If we allow group convolution to obtain input data from different groups (as shown in Fig~\ref{fig:channelshuffle} (b)), the input and output channels will be fully related. Specifically, for the feature map generated from the previous group layer, we can first divide the channels in each group into several subgroups, then feed each group in the next layer with different subgroups. This can be efficiently and elegantly implemented by a \emph{channel shuffle} operation (Fig~\ref{fig:channelshuffle} (c)): suppose a convolutional layer with $g$ groups whose output has $g\times n$ channels; we first reshape the output channel dimension into $(g, n)$, transposing and then flattening it back as the input of next layer. Note that the operation still takes effect even if the two convolutions have different numbers of groups. Moreover, channel shuffle is also differentiable, which means it can be embedded into network structures for end-to-end training.

Channel shuffle operation makes it possible to build more powerful structures with multiple group convolutional layers. In the next subsection we will introduce an efficient network unit with channel shuffle and group convolution.

\subsection{ShuffleNet Unit}
\label{sec:unit}

\begin{table*}[ht]
	\begin{center}
		\begin{tabular}{l|c|c|c|c|ccccc}
			\hline
			Layer & Output size & KSize & Stride & Repeat & \multicolumn{5}{c}{Output channels ($g$ groups)} \\
			%\cmidrule{6-10}
			%\hline
			& & & & & $g=1$ & $g=2$ & $g=3$ & $g=4$ & $g=8$ \\
			\hline \hline
			Image & $224\times 224$ & & & & 3 & 3 & 3 & 3 & 3 \\
			\hline
			Conv1 & $112\times 112$ & $3\times 3$ & 2 & 1 & 24 & 24 & 24 & 24 & 24 \\
			MaxPool & $56\times 56$ & $3\times 3$ & 2 &  &  &  &  &  &  \\
			\hline
			Stage2 & $28\times 28$ & & 2 & 1 & 144 & 200 & 240 & 272 & 384 \\
			& $28\times 28$ & & 1 & 3 & 144 & 200 & 240 & 272 & 384 \\
			\hline
			Stage3 & $14\times 14$ & & 2 & 1 & 288 & 400 & 480 & 544 & 768 \\
			& $14\times 14$ & & 1 & 7 & 288 & 400 & 480 & 544 & 768 \\
			\hline
			Stage4 & $7\times 7$ & & 2 & 1 & 576 & 800 & 960 & 1088 & 1536 \\
			& $7\times 7$ & & 1 & 3 & 576 & 800 & 960 & 1088 & 1536 \\
			\hline
			GlobalPool & $1\times1$ & $7\times 7$ & & & & & & & \\ 
			\hline
			FC & & & & & 1000 & 1000 & 1000 & 1000 & 1000 \\
			\hline \hline
			Complexity & & & & & 143M & 140M & 137M & 133M & 137M \\
			\hline
		\end{tabular}
	\end{center}
	\caption{ShuffleNet architecture. The complexity is evaluated with FLOPs, i.e. the number of floating-point multiplication-adds. Note that for Stage 2, we do not apply group convolution on the first pointwise layer because the number of input channels is relatively small.}
	\label{tbl:arch}
\end{table*}

\begin{table*}[t]
	\begin{center}
		\begin{tabular}{l|c|ccccc}
			\hline
			Model & Complexity & \multicolumn{5}{c}{Classification error (\%)} \\
			%\cmidrule{3-7}
			& (MFLOPs) & $g=1$ & $g=2$ & $g=3$ & $g=4$ & $g=8$ \\
			\hline
			\hline
			ShuffleNet 1$\times$ & 140 & 33.6 & 32.7 & 32.6 & 32.8 & \textbf{32.4} \\
			ShuffleNet 0.5$\times$ & 38	& 45.1 & 44.4 & 43.2 & \textbf{41.6} & 42.3 \\
			ShuffleNet 0.25$\times$ & 13 & 57.1 & 56.8 & 55.0  & 54.2 & \textbf{52.7} \\
			\hline
		\end{tabular}
	\end{center}
	\caption{Classification error vs. number of groups $g$ (\emph{smaller number represents better performance})}
	\label{tbl:groupconv}
\end{table*}

Taking advantage of the channel shuffle operation, we propose a novel \emph{ShuffleNet} unit specially designed for small networks. We start from the design principle of bottleneck unit~\cite{he2016deep} in Fig~\ref{fig:unit} (a). It is a residual block. In its residual branch, for the $3\times 3$ layer, we apply a computational economical $3\times 3$ depthwise convolution~\cite{chollet2016xception} on the bottleneck feature map. Then, we replace the first $1\times 1$ layer with pointwise group convolution followed by a channel shuffle operation, to form a ShuffleNet unit, as shown in Fig~\ref{fig:unit} (b). The purpose of the second pointwise group convolution is to recover the channel dimension to match the shortcut path. For simplicity, we do not apply an extra channel shuffle operation after the second pointwise layer as it results in comparable scores. The usage of batch normalization (BN)~\cite{ioffe2015batch} and nonlinearity is similar to ~\cite{he2016deep,xie2016aggregated}, except that we do not use ReLU after depthwise convolution as suggested by ~\cite{chollet2016xception}. As for the case where ShuffleNet is applied with stride, we simply make two modifications (see Fig~\ref{fig:unit} (c)): (i) add a $3\times 3$ average pooling on the shortcut path; (ii) replace the element-wise addition with channel concatenation, which makes it easy to enlarge channel dimension with little extra computation cost.

Thanks to pointwise group convolution with channel shuffle, all components in ShuffleNet unit can be computed efficiently. Compared with ResNet~\cite{he2016deep} (bottleneck design) and ResNeXt~\cite{xie2016aggregated}, our structure has less complexity under the same settings. For example, given the input size $c\times h \times w$ and the bottleneck channels $m$, ResNet unit requires $hw(2cm+9m^2)$ FLOPs and ResNeXt has $hw(2cm+9m^2/g)$ FLOPs, while our ShuffleNet unit requires only $hw(2cm/g+9m)$ FLOPs, where $g$ means the number of groups for convolutions. In other words, given a computational budget, ShuffleNet can use wider feature maps. We find this is critical for small networks, as tiny networks usually have an insufficient number of channels to process the information.

In addition, in ShuffleNet depthwise convolution only performs on bottleneck feature maps. Even though depthwise convolution usually has very low theoretical complexity, we find it difficult to efficiently implement on low-power mobile devices, which may result from a worse computation/memory access ratio compared with other dense operations. Such drawback is also referred in ~\cite{chollet2016xception}, which has a runtime library based on TensorFlow~\cite{abadi2016tensorflow}. In ShuffleNet units, we intentionally use depthwise convolution only on bottleneck in order to prevent overhead as much as possible.

\subsection{Network Architecture}

Built on ShuffleNet units, we present the overall ShuffleNet architecture in Table \ref{tbl:arch}. The proposed network is mainly composed of a stack of ShuffleNet units grouped into three stages. The first building block in each stage is applied with stride = 2. Other hyper-parameters within a stage stay the same, and for the next stage the output channels are doubled. Similar to ~\cite{he2016deep}, we set the number of bottleneck channels to 1/4 of the output channels for each ShuffleNet unit. Our intent is to provide a reference design as simple as possible, although we find that further hyper-parameter tunning might generate better results.

In ShuffleNet units, group number $g$ controls the connection sparsity of pointwise convolutions. Table~\ref{tbl:arch} explores different group numbers and we adapt the output channels to ensure overall computation cost roughly unchanged ($\sim$140 MFLOPs). Obviously, larger group numbers result in more output channels (thus more convolutional filters) for a given complexity constraint, which helps to encode more information, though it might also lead to degradation for an individual convolutional filter due to limited corresponding input channels. In Sec~\ref{sec:gconv} we will study the impact of this number subject to different computational constrains.

To customize the network to a desired complexity, we can simply apply a scale factor $s$ on the number of channels. For example, we denote the networks in Table~\ref{tbl:arch} as "ShuffleNet 1$\times$", then "ShuffleNet $s\times$" means scaling the number of filters in ShuffleNet 1$\times$ by $s$ times thus overall complexity will be roughly $s^2$ times of ShuffleNet 1$\times$. 

\section{Experiments}
\label{sec:exp}

\begin{table*}[ht]
	\begin{center}
		\begin{tabular}{l|c|c|c}
			\hline
			Model & Cls err. (\%, no shuffle) & Cls err. (\%, shuffle) & $\Delta$ err. (\%) \\
			\hline \hline
			ShuffleNet 1x ($g=3$) & 34.5 & \textbf{32.6} & 1.9 \\
			ShuffleNet 1x ($g=8$) & 37.6 & \textbf{32.4} & 5.2 \\
			\hline
			ShuffleNet 0.5x ($g=3$) & 45.7 & \textbf{43.2} & 2.5 \\
			ShuffleNet 0.5x ($g=8$) & 48.1 & \textbf{42.3} & 5.8 \\
			\hline
			ShuffleNet 0.25x ($g=3$) & 56.3 & \textbf{55.0} & 1.3 \\
			ShuffleNet 0.25x ($g=8$) & 56.5 & \textbf{52.7} & 3.8 \\
			\hline
		\end{tabular}
	\end{center}
	\caption{ShuffleNet with/without channel shuffle (\emph{smaller number represents better performance})}
	\label{tbl:shuffle}
\end{table*}

We mainly evaluate our models on the ImageNet 2012 classification dataset~\cite{russakovsky2015imagenet,deng2009imagenet}. We follow most of the training settings and hyper-parameters used in ~\cite{xie2016aggregated}, with two exceptions: (i) we set the weight decay to 4e-5 instead of 1e-4 and use linear-decay learning rate policy (decreased from 0.5 to 0); (ii) we use slightly less aggressive scale augmentation for data preprocessing. Similar modifications are also referenced in ~\cite{howard2017mobilenets} because such small networks usually suffer from underfitting rather than overfitting. It takes 1 or 2 days to train a model for $3\times 10^5$ iterations on 4 GPUs, whose batch size is set to 1024. To benchmark, we compare single crop top-1 performance on ImageNet validation set, i.e. cropping $224\times 224$ center view from $256\times$ input image and evaluating classification accuracy. We use exactly the same settings for all models to ensure fair comparisons.

\subsection{Ablation Study}

The core idea of ShuffleNet lies in pointwise group convolution and channel shuffle operation. In this subsection we evaluate them respectively.

\subsubsection{Pointwise Group Convolutions}
\label{sec:gconv}

To evaluate the importance of pointwise group convolutions, we compare ShuffleNet models of the same complexity whose numbers of groups range from 1 to 8. If the group number equals 1, no pointwise group convolution is involved and then the ShuffleNet unit becomes an "Xception-like"~\cite{chollet2016xception} structure. For better understanding, we also scale the width of the networks to 3 different complexities and compare their classification performance respectively. Results are shown in Table~\ref{tbl:groupconv}.

From the results, we see that models with group convolutions ($g>1$) consistently perform better than the counterparts without pointwise group convolutions ($g=1$). Smaller models tend to benefit more from groups. For example, for ShuffleNet 1$\times$ the best entry ($g=8$) is 1.2\% better than the counterpart, while for ShuffleNet 0.5$\times$ and 0.25$\times$ the gaps become 3.5\% and 4.4\% respectively. Note that group convolution allows more feature map channels for a given complexity constraint, so we hypothesize that the performance gain comes from wider feature maps which help to encode more information. In addition, a smaller network involves thinner feature maps, meaning it benefits more from enlarged feature maps.

Table \ref{tbl:groupconv} also shows that for some models (e.g. ShuffleNet 0.5$\times$) when group numbers become relatively large (e.g. $g=8$), the classification score saturates or even drops. With an increase in group number (thus wider feature maps), input channels for each convolutional filter become fewer, which may harm representation capability. Interestingly, we also notice that for smaller models such as ShuffleNet 0.25$\times$ larger group numbers tend to better results consistently, which suggests wider feature maps bring more benefits for smaller models. 

\subsubsection{Channel Shuffle vs. No Shuffle}
\label{sec:shuffle}

The purpose of shuffle operation is to enable cross-group information flow for multiple group convolution layers. Table~\ref{tbl:shuffle} compares the performance of ShuffleNet structures (group number is set to 3 or 8 for instance) with/without channel shuffle. The evaluations are performed under three different scales of complexity. It is clear that channel shuffle consistently boosts classification scores for different settings. Especially, when group number is relatively large (e.g. $g=8$), models with channel shuffle outperform the counterparts by a significant margin, which shows the importance of cross-group information interchange.

\subsection{Comparison with Other Structure Units}
\label{sec:compother}

\begin{table*}[ht]
	\begin{center}
		\begin{tabular}{c|c|c|c|c|l}
			\hline
			Complexity (MFLOPs) & VGG-like & ResNet & Xception-like & ResNeXt & ShuffleNet (ours) \\
			\hline
			\hline
			140 & 50.7 & 37.3 & 33.6 & 33.3 & \textbf{32.4} (1$\times$, $g=8$) \\
			38 & - & 48.8 & 45.1 & 46.0 & \textbf{41.6} (0.5$\times$, $g=4$) \\
			13 & - & 63.7 & 57.1 & 65.2 & \textbf{52.7} (0.25$\times$, $g=8$) \\
			\hline
		\end{tabular}
	\end{center}
	\caption{Classification error vs. various structures (\%, \emph{smaller number represents better performance}). We do not report VGG-like structure on smaller networks because the accuracy is significantly worse.}
	\label{tbl:structues}
\end{table*}

\begin{table*}[h]
	\begin{center}
		\begin{tabular}{l|c|c|c}
			\hline
			Model & Complexity (MFLOPs) & Cls err. (\%) & $\Delta$ err. (\%) \\
			\hline\hline
			1.0 MobileNet-224 & 569 & 29.4 & - \\
			ShuffleNet 2$\times$ ($g=3$) & 524 & \textbf{26.3} & 3.1 \\
			ShuffleNet 2$\times$ (with \emph{SE}\cite{hu2017squeeze}, $g=3$) & 527 & \textbf{24.7} & 4.7 \\
			\hline
			0.75 MobileNet-224 & 325 & 31.6 & - \\
			ShuffleNet 1.5$\times$ ($g=3$) & 292 & \textbf{28.5} & 3.1 \\
			\hline
			0.5 MobileNet-224 & 149 & 36.3 & - \\
			ShuffleNet 1$\times$ ($g=8$) & 140 & \textbf{32.4} & 3.9 \\
			\hline
			0.25 MobileNet-224 & 41 & 49.4 & - \\
			ShuffleNet 0.5$\times$ ($g=4$) & 38 & \textbf{41.6} & 7.8 \\
			ShuffleNet 0.5$\times$ (shallow, $g=3$) & 40 & 42.8 & 6.6 \\
			\hline
		\end{tabular}
	\end{center}
	\caption{ ShuffleNet vs. MobileNet~\cite{howard2017mobilenets} on ImageNet Classification }%(\emph{smaller number represents better performance})}
	\label{tbl:mobilecls}
\end{table*}

\begin{table*}[h]
	\begin{center}
		\begin{tabular}{l|c|c}
			\hline
			Model & Cls err. (\%) & Complexity (MFLOPs) \\
			\hline\hline
			VGG-16~\cite{simonyan2014very} & 28.5 & 15300 \\
			ShuffleNet 2$\times$ ($g=3$) & 26.3 & \textbf{524} \\
			\hline
			GoogleNet~\cite{szegedy2015going}* & 31.3 & 1500 \\
			ShuffleNet 1$\times$ ($g=8$) & 32.4 & \textbf{140} \\
			\hline
			AlexNet~\cite{krizhevsky2012imagenet} & 42.8 & 720 \\
			SqueezeNet~\cite{iandola2016squeezenet} & 42.5 & 833 \\
			ShuffleNet 0.5$\times$ ($g=4$) & 41.6 & \textbf{38} \\
			\hline 
		\end{tabular}
	\end{center}
	\caption{ Complexity comparison. *\emph{Implemented by BVLC (https://github.com/BVLC/caffe/tree/master/models/bvlc\_googlenet)} }
	\label{tbl:complexity}
\end{table*}

Recent leading convolutional units in VGG~\cite{simonyan2014very}, ResNet~\cite{he2016deep}, GoogleNet~\cite{szegedy2015going}, ResNeXt~\cite{xie2016aggregated} and Xception~\cite{chollet2016xception} have pursued state-of-the-art results with large models (e.g. $\ge 1$GFLOPs), but do not fully explore low-complexity conditions. In this section we survey a variety of building blocks and make comparisons with ShuffleNet under the same complexity constraint.

For fair comparison, we use the overall network architecture as shown in Table~\ref{tbl:arch}. We replace the ShuffleNet units in Stage 2-4 with other structures, then adapt the number of channels to ensure the complexity remains unchanged. The structures we explored include:

\begin{itemize}

	\item \emph{VGG-like}. Following the design principle of VGG net~\cite{simonyan2014very}, we use a two-layer 3$\times$3 convolutions as the basic building block. Different from ~\cite{simonyan2014very}, we add a Batch Normalization layer~\cite{ioffe2015batch} after each of the convolutions to make end-to-end training easier.
		
	\item \emph{ResNet}. We adopt the "bottleneck" design in our experiment, which has been demonstrated more efficient in ~\cite{he2016deep} . Same as ~\cite{he2016deep}, the \emph{bottleneck ratio}\footnote{In the bottleneck-like units (like ResNet, ResNeXt or ShuffleNet) \emph{bottleneck ratio} implies the ratio of bottleneck channels to output channels. For example, bottleneck ratio = $1:4$ means the output feature map is 4 times the width of the bottleneck feature map. } is also $1:4$. 
	
	\item \emph{Xception-like}. The original structure proposed in ~\cite{chollet2016xception} involves fancy designs or hyper-parameters for different stages, which we find difficult for fair comparison on small models. Instead, we remove the pointwise group convolutions and channel shuffle operation from ShuffleNet (also equivalent to ShuffleNet with $g=1$). The derived structure shares the same idea of ``depthwise separable convolution" as in ~\cite{chollet2016xception}, which is called an \emph{Xception-like} structure here.
	
	\item \emph{ResNeXt}. We use the settings of \emph{cardinality} $=16$ and bottleneck ratio $=1:2$ as suggested in ~\cite{xie2016aggregated}. We also explore other settings, e.g. bottleneck ratio $=1:4$, and get similar results.
	
\end{itemize}

\begin{table*}[h]
	\begin{center}
		\begin{tabular}{l|c|c}
			\hline
			Model & mAP $[.5,.95]$ (300$\times$ image) &mAP $[.5,.95]$ (600$\times$ image) \\
			\hline\hline
			ShuffleNet 2$\times$ ($g=3$) & \textbf{18.7\%} & \textbf{25.0\%} \\
			ShuffleNet 1$\times$ ($g=3$) & 14.5\% & 19.8\% \\
			\hline
			1.0 MobileNet-224~\cite{howard2017mobilenets} & 16.4\% & 19.8\% \\
			1.0 MobileNet-224 (our impl.) & 14.9\% & 19.3\% \\
			\hline 
		\end{tabular}
	\end{center}
	\caption{ Object detection results on MS COCO (\emph{larger numbers represents better performance}). For MobileNets we compare two results: 1) COCO detection scores reported by \cite{howard2017mobilenets}; 2) finetuning from our reimplemented MobileNets, whose training and finetuning settings are exactly the same as that for ShuffleNets. }
	\label{tbl:det}
\end{table*}

\begin{table*}[!ht]
	\begin{center}
		\begin{tabular}{l|c|c|c|c|c}
			\hline
			Model & Cls err. (\%) & FLOPs & $224\times 224$ & $480\times 640$ & $720\times 1280$ \\
			\hline\hline
			ShuffleNet 0.5$\times$ ($g=3$) & 43.2 & 38M & 15.2ms & 87.4ms & 260.1ms \\
			ShuffleNet 1$\times$ ($g=3$) & 32.6 & 140M & 37.8ms & 222.2ms & 684.5ms \\
			ShuffleNet 2$\times$ ($g=3$) & 26.3 & 524M & 108.8ms & 617.0ms & 1857.6ms \\
			\hline
			AlexNet~\cite{krizhevsky2012imagenet} & 42.8 & 720M & 184.0ms & 1156.7ms & 3633.9ms \\
			1.0 MobileNet-224~\cite{howard2017mobilenets} & 29.4 & 569M & 110.0ms & 612.0ms & 1879.2ms \\
			\hline 
		\end{tabular}
	\end{center}
	\caption{ Actual inference time on mobile device (\emph{smaller number represents better performance}). The platform is based on a single Qualcomm Snapdragon 820 processor. All results are evaluated with \textbf{single thread}. }
	\label{tbl:actual}
\end{table*}

We use exactly the same settings to train these models. Results are shown in Table~\ref{tbl:structues}. Our ShuffleNet models outperform most others by a significant margin under different complexities. Interestingly, we find an empirical relationship between feature map channels and classification accuracy. For example, under the complexity of 38 MFLOPs, output channels of Stage 4 (see Table~\ref{tbl:arch}) for VGG-like, ResNet, ResNeXt, Xception-like, ShuffleNet models are 50, 192, 192, 288, 576 respectively, which is consistent with the increase of accuracy. Since the efficient design of ShuffleNet, we can use more channels for a given computation budget, thus usually resulting in better performance.

Note that the above comparisons do not include \emph{GoogleNet} or \emph{Inception} series~\cite{szegedy2015going,szegedy2016rethinking,szegedy2016inception}. We find it nontrivial to generate such Inception structures to small networks because the original design of Inception module involves too many hyper-parameters. As a reference, the first GoogleNet version~\cite{szegedy2015going} has 31.3\% top-1 error at the cost of 1.5 GFLOPs (See Table~\ref{tbl:complexity}). More sophisticated Inception versions~\cite{szegedy2016rethinking,szegedy2016inception} are more accurate, however, involve significantly increased complexity. Recently, Kim et al. propose a lightweight network structure named \emph{PVANET}~\cite{kim2016pvanet} which adopts Inception units. Our reimplemented PVANET (with 224$\times$224 input size) has 29.7\% classification error with a computation complexity of 557 MFLOPs, while our ShuffleNet 2x model ($g=3$) gets 26.3\% with 524 MFLOPs (see Table~\ref{tbl:complexity}). 
\subsection{Comparison with MobileNets and Other Frameworks}
\label{sec:compmobile}

Recently Howard et al. have proposed \emph{MobileNets}~\cite{howard2017mobilenets} which mainly focus on efficient network architecture for mobile devices. MobileNet takes the idea of depthwise separable convolution from ~\cite{chollet2016xception} and achieves state-of-the-art results on small models.

Table~\ref{tbl:mobilecls} compares classification scores under a variety of complexity levels. It is clear that our ShuffleNet models are superior to MobileNet for all the complexities. Though our ShuffleNet network is specially designed for small models ($<150$ MFLOPs), we find it is still better than MobileNet for higher computation cost, e.g. 3.1\% more accurate than MobileNet 1$\times$ at the cost of 500 MFLOPs. For smaller networks ($\sim$40 MFLOPs) ShuffleNet surpasses MobileNet by \textbf{7.8\%}. Note that our ShuffleNet architecture contains 50 layers while MobileNet only has 28 layers. For better understanding, we also try ShuffleNet on a 26-layer architecture by removing half of the blocks in Stage 2-4 (see "ShuffleNet 0.5$\times$ shallow ($g=3$)" in Table~\ref{tbl:mobilecls}). Results show that the shallower model is still significantly better than the corresponding MobileNet, which implies that the effectiveness of ShuffleNet mainly results from its efficient structure, not the depth.

Table~\ref{tbl:complexity} compares our ShuffleNet with a few popular models. Results show that with similar accuracy ShuffleNet is much more efficient than others. For example, ShuffleNet 0.5$\times$ is theoretically 18$\times$ faster than AlexNet~\cite{krizhevsky2012imagenet} with comparable classification score. We will evaluate the actual running time in Sec \ref{sec:actual}.

It is also worth noting that the simple architecture design makes it easy to equip ShuffeNets with the latest advances such as \cite{hu2017squeeze, ramachandran2017swish}. For example, in \cite{hu2017squeeze} the authors propose \emph{Squeeze-and-Excitation (SE)} blocks which achieve state-of-the-art results on large ImageNet models. We find SE modules also take effect in combination with the backbone ShuffleNets, for instance, boosting the top-1 error of ShuffleNet 2$\times$ to 24.7\% (shown in Table~\ref{tbl:mobilecls}). Interestingly, though negligible increase of theoretical complexity, we find ShuffleNets with SE modules are usually $25 \sim 40\%$ slower than the ``raw'' ShuffleNets on mobile devices, which implies that actual speedup evaluation is critical on low-cost architecture design. In Sec~\ref{sec:actual} we will make further discussion.

\subsection{Generalization Ability}
To evaluate the generalization ability for transfer learning, we test our ShuffleNet model on the task of MS COCO object detection~\cite{lin2014microsoft}. We adopt Faster-RCNN~\cite{ren2015faster} as the detection framework and use the publicly released Caffe code~\cite{ren2015faster,jia2014caffe} for training with default settings. Similar to ~\cite{howard2017mobilenets}, the models are trained on the COCO train+val dataset excluding 5000 minival images and we conduct testing on the minival set. Table~\ref{tbl:det} shows the comparison of results trained and evaluated on two input resolutions. Comparing ShuffleNet 2$\times$ with MobileNet whose complexity are comparable (524 vs. 569 MFLOPs), our ShuffleNet 2$\times$ surpasses MobileNet by a significant margin on both resolutions; our ShuffleNet 1$\times$ also achieves comparable results with MobileNet on 600$\times$ resolution, but has $\sim$4$\times$ complexity reduction. We conjecture that this significant gain is partly due to ShuffleNet's simple design of architecture without bells and whistles.

\subsection{Actual Speedup Evaluation}
\label{sec:actual}

Finally, we evaluate the actual inference speed of ShuffleNet models on a mobile device with an ARM platform. Though ShuffleNets with larger group numbers (e.g. $g=4$ or $g=8$) usually have better performance, we find it less efficient in our current implementation. Empirically $g=3$ usually has a proper trade-off between accuracy and actual inference time. As shown in Table~\ref{tbl:actual}, three input resolutions are exploited for the test. Due to memory access and other overheads, we find every 4$\times$ theoretical complexity reduction usually results in $\sim$2.6$\times$ actual speedup in our implementation. Nevertheless, compared with AlexNet~\cite{krizhevsky2012imagenet} our ShuffleNet 0.5$\times$ model still achieves $\sim$13$\times$ \textbf{actual} speedup under comparable classification accuracy (the theoretical speedup is 18$\times$), which is much faster than previous AlexNet-level models or speedup approaches such as ~\cite{iandola2016squeezenet,jaderberg2014speeding,lebedev2014speeding,zhang2016accelerating,zhang2015efficient,wen2016learning}.

%\section{Conclusion}
%Deploying neural networks on mobile devices is of great importance. To this end, we utilize pointwise group convolution and channel shuffle operation to devise a new CNN architecture called ShuffleNet that operates at low computation cost while remains accurate. Note that, our approach may be orthogonal to other acceleration methods like quantization~\cite{zhou2017incremental}, atnd pruning~\cite{han2015deep}. Utilizing these techniques may further accelerate the model.

{\small
\bibliographystyle{ieee}
\bibliography{shufflenet}
}

\end{document}